%% file: iclr2024_conference.tex
\documentclass{article} 
\usepackage{iclr2024_conference,times}

\input{math_commands.tex}

\usepackage{hyperref}
\usepackage{url}
\usepackage{cite}
\usepackage{amsmath,amssymb,amsfonts}
\usepackage{algorithmic}
\usepackage{graphicx}
\usepackage{textcomp}
\usepackage{xcolor}
\usepackage{subcaption}
\usepackage{wrapfig}
\graphicspath{ {./images/} }

\title{Neural Tree Reconstruction for the \\ Open Forest Observatory}

\author{Marissa Ramirez de Chanlatte, Arjun Rewari \& Trevor Darrell \\
Berkeley AI Research\\
University of California, Berkeley\\
\texttt{\{marissachanlatte, arjun.rewari, trevordarrell\}@berkeley.edu} \\
\And
Derek J. N. Young \\
Department of Plant Sciences \\
University of California, Davis \\
\texttt{djyoung@ucdavis.edu} \\
}

\iclrfinalcopy 
\begin{document}

\maketitle

\begin{abstract}
The Open Forest Observatory (OFO) is a collaboration across universities and other partners to make low-cost forest mapping accessible to ecologists, land managers, and the general public. The OFO is building both a database of geospatial forest data as well as open-source methods and tools for forest mapping by uncrewed aerial vehicle. Such data are useful for a variety of climate applications including prioritizing reforestation efforts, informing wildfire hazard reduction, and monitoring carbon sequestration. In the current iteration of the OFO's forest map database, 3D tree maps are created using classical structure-from-motion techniques. This approach is prone to artifacts, lacks detail, and has particular difficulty on the forest floor where the input data (overhead imagery) has limited visibility. These reconstruction errors can potentially propagate to the downstream scientific tasks (e.g. a wildfire simulation.) Advances in 3D reconstruction, including methods like Neural Radiance Fields (NeRF), produce higher quality results that are more robust to sparse views and support data-driven priors. We explore ways to incorporate NeRFs into the OFO dataset, outline future work to support even more state-of-the-art 3D vision models, and describe the importance of high-quality 3D reconstructions for forestry applications.
\end{abstract}

\section{Introduction}

\begin{wrapfigure}{r}{.5\textwidth}
\vspace{-1em}
\centering
\includegraphics[width=\linewidth]{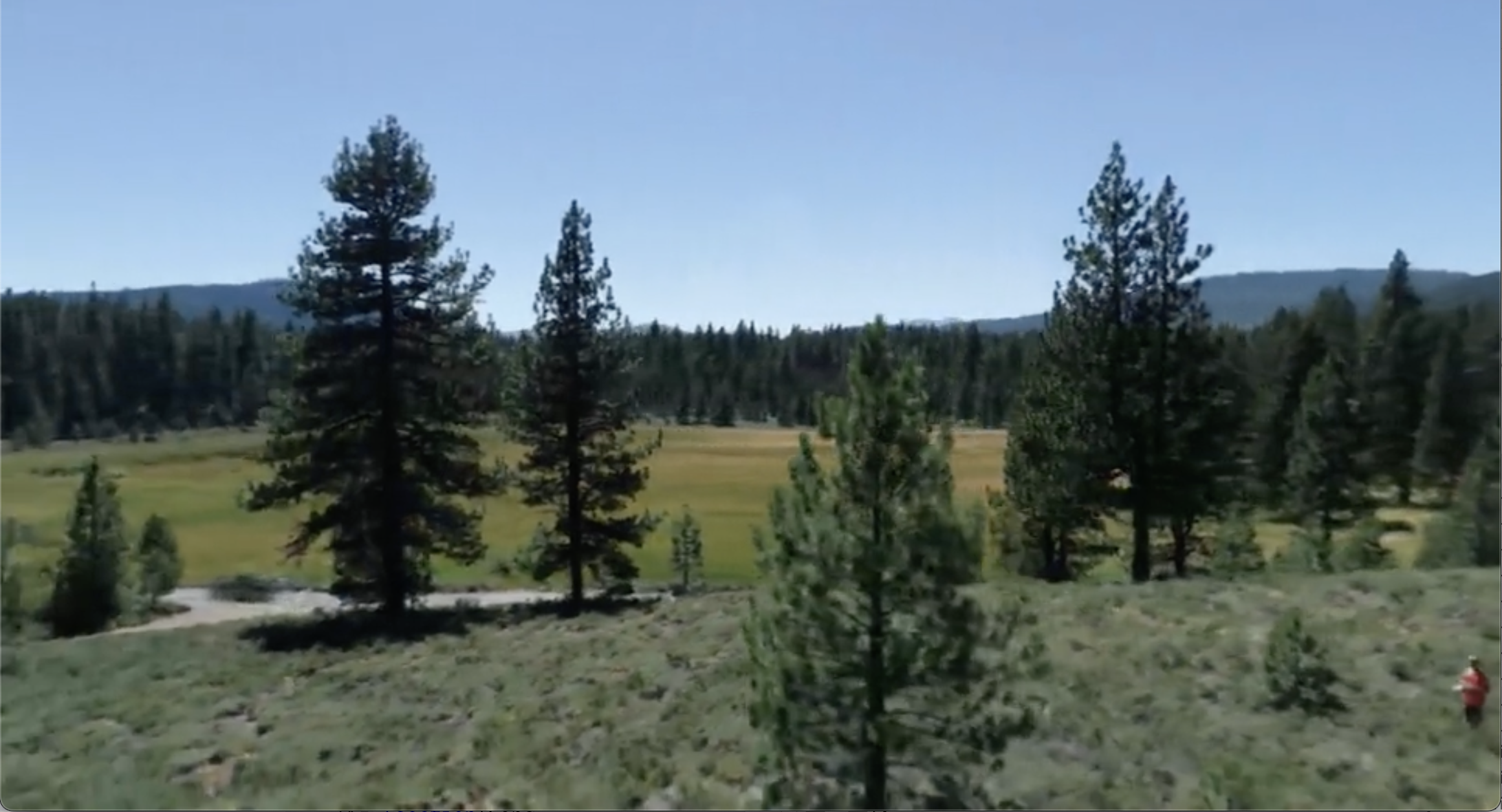}
    \caption{A rendered view of a NeRF created from drone-captured forest imagery.}
    \label{nerf}
\end{wrapfigure}
The Open Forest Observatory (OFO) is an open source software and data repository created in a joint effort by UC Davis, University of Arizona, and CU Boulder to facilitate drone-based forest mapping for forest ecology and management applications. The dataset in development includes drone-derived 2D photos and videos of forests, photogrammetry-derived 2D and 3D forest maps~\citep{young2022optimizing}, and field-based tree geolocation and species classification information, among other forest data. 


Currently, the 3D forest models in the OFO are created using structure-from-motion (SfM) ~\citep{schonberger2016structure}, a classical photogrammetry technique. SfM uses keypoints in images to reconstruct a point-cloud and depth maps that can then be processed into a 3D mesh. It is not designed to synthesize photo-realistic novel views in the absence of complete data and the meshing approaches often obscure fine detail that can be useful in semantic classification.

Several recent advances in 3D reconstruction have been made to ameliorate these problems, however little study has been made on applications to forest data, especially for scientific applications, which have different thresholds for acceptable error. The OFO has partnered with computer vision researchers to explore translating state-of-the-art computer vision models to this vital application area. This effort is on-going and in this paper we report on its progress and introduce the OFO dataset to the AI community to encourage further collaboration.

Specifically we explore Neural Radiance Fields (NeRFs) ~\citep{mildenhall2021nerf} as applied to forest mapping. We use NeRF to improve the reconstruction quality in the OFO. NeRFs have been shown to have much higher reconstruction quality and are able to fill in novel views~\citep{yu2021pixelnerf}. In our experiments, we show that NeRF is not only better at creating a more photorealisitic 3D scene, it also produces a higher quality mesh.  Photorealism has benefits beyond aesthetics. Understanding density and biomass in a forest are essential to tasks such as estimation of fuel for wildfires and carbon sequestration. While estimating these quantities requires some 3D understanding of the world, it does not necessarily require an exact 3D geometry. There are many qualitative measurements that are currently only done by sending an expert to physically walk through a forest to estimate things like fuel density or tree spacing. A sufficiently high quality visual experience could facilitate virtual remote qualitative assessments by experts. Such renderings could even enable neural networks to eventually augment or conceivably replace these human assessments. Further validation and bench-marking could allow for quantitative assessments as well, for example counting trees and estimating volumes. 

\textbf{Contributions:} In this paper, we introduce the Open Forest Observatory, demonstrate qualitative visual improvement (e.g. clearer tree structure, trunks, and leaves) with NeRF on forest data leading to greater opportunity for virtual qualitative assessment, and outline the various extensions of this technology that can further transform the way forests are managed. 

\section{Background}
\textbf{Forest Mapping:} Forest inventories are critical resources for understanding ecological processes and informing forest management, but they have traditionally required time-consuming ground-based surveys. Individual tree-level measurements including location, height, stem diameter, health status, and species identity require substantial time and effort to obtain. 
Advances in drone and imagery processing technology are enabling a new era of forest research in which individual trees can be mapped, measured, and identified to genus or species across broad areas without extensive ground surveys ~\citep{mlambo2017structure,camarretta2020monitoring}, reducing survey time to days rather than months. Drone-based mapping approaches generally involve executing a drone mission to collect a series of images with high (70-90\%) image-to-image overlap over a contiguous study area. The images are then traditionally processed using photogrammetry into downstream data products such as an orthomosaic, canopy height model, and a 3D mesh model of the scene. In turn, one or more of these products can be used extract forest inventory data; for example, individual treetops may be detected as local maxima in the canopy height model ~\citep{young2022optimizing} and trees may be classified to species based on their appearance in the orthomosaic ~\citep{ferreira2020individual, weinstein2023capturing}. 


\textbf{Neural Radiance Fields (NeRF):} NeRF~\citep{mildenhall2021nerf} is a neural network based technique for 3D reconstruction. It reconstructs accurate 3D scenes by modeling the volumetric scene geometry and appearance from a set of images, and is designed to synthesize novel views of complex scenes. A NeRF is an implicit function that takes in a set of 3D points \textit{(x, y, z)} and viewing directions ($\theta, \phi$) from a scene which are then fed through a multilayer perceptron (MLP) to produce a pixel value with a predicted set of colors \textit{(r, g, b)} and density \textit{$\sigma$}. To create a novel view of the scene, NeRF employs ray casting on each pixel in the view, creating a set of colors and densities along ray at random points. Classical volume rendering techniques are then used to accumulate those colors and densities into a 2D image using the following color accumulation equation: $\hat{\textbf{c}}(\textbf{o, d}) = \sum\limits_{i=1}^N w_i \textbf{c}_i, $ where the integration is weighted by $w_i = T_i\alpha_i$, the accumulated transmittance (the fraction of light that reaches the camera) times the opacity of the i-th ray segment. To supervise the neural network, the predicted value is then compared with the true color $\textbf{c}$ of the input image via the following loss function: $\mathcal{L}_{RGB} = ||\hat{\textbf{c}} - \textbf{c}||_1.$

\section{Neural Reconstruction Proof of Concept}

The OFO is a program created by forest ecologists for other ecologists, land managers, and the general public. The OFO aims to integrate existing and emerging imagery processing tools and techniques, tune them for forestry applications, and present them through user-friendly interfaces. The OFO's initial focus on classical photogrammetry-based approaches has achieved high-quality mapping of the forest overstory~\citep{young2022optimizing}, but the limited utility of these methods for understory reconstruction poses some constraints to widespread adoption by the forest ecology and management community. In particular, most forest inventory methods, management prescriptions, and ecological analyses rely critically on measurements of tree stem diameter at breast height (DBH), which can be challenging to obtain using photogrammetry even in sparse stands ~\citep{swayze2021influence, swayze2022application}.

\begin{wrapfigure}[15]{l}{0.5\textwidth}
\vspace{-1em}
\centering
\includegraphics[width=.4\textwidth]{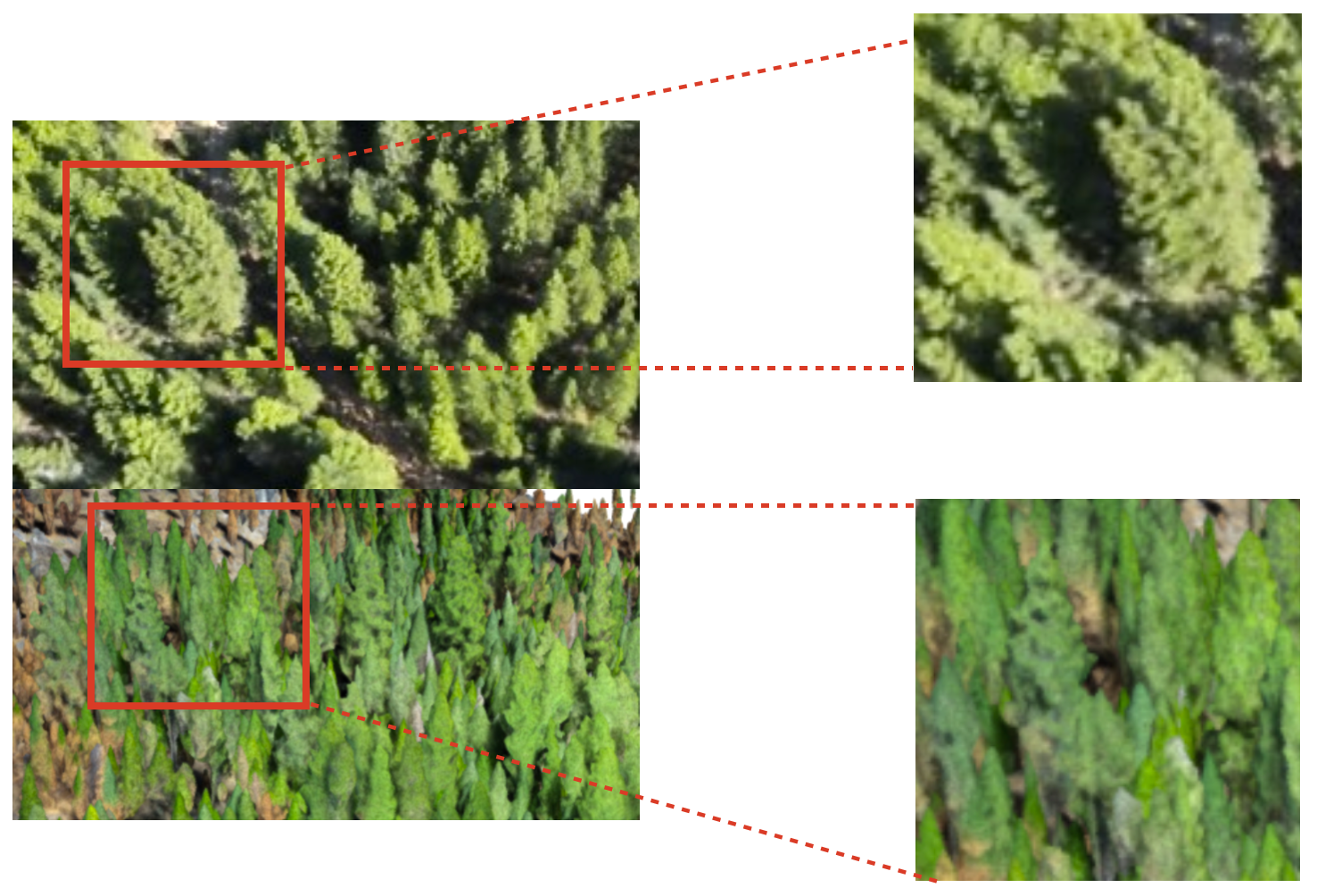}

\caption{NeRF (top) versus current SfM mesh reconstruction (below) on overhead drone imagery. The NeRF retains more fine detail in the trees, most notably leaf and branch structure.}
\label{fig:teaser}
\end{wrapfigure}

The project described in this paper represents a recent effort to bring in computer vision researchers to the OFO to explore translating state-of-the-art models for 3D reconstruction to real forestry applications. This work is on-going, but we share our progress in an effort to encourage others in the AI community to engage with this dataset and contribute to this vital application area.

\begin{wrapfigure}[18]{r}{.4\textwidth}
\vspace{-2em}
\centering
\begin{subfigure}[b]{0.3\textwidth}
   \includegraphics[width=1\linewidth]{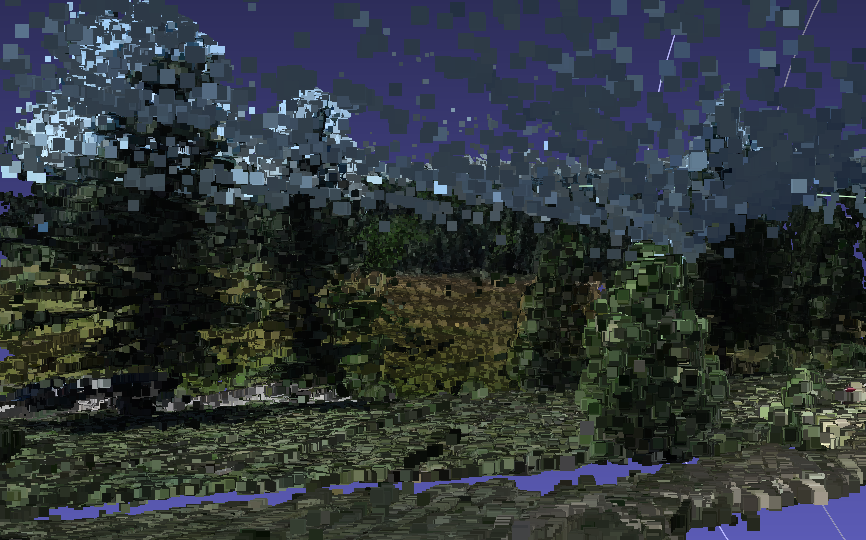}
   \caption{Point-cloud generated via SfM.}
   \label{fig:Ng1} 
\end{subfigure}
\begin{subfigure}[b]{0.3\textwidth}
   \includegraphics[width=1\linewidth]{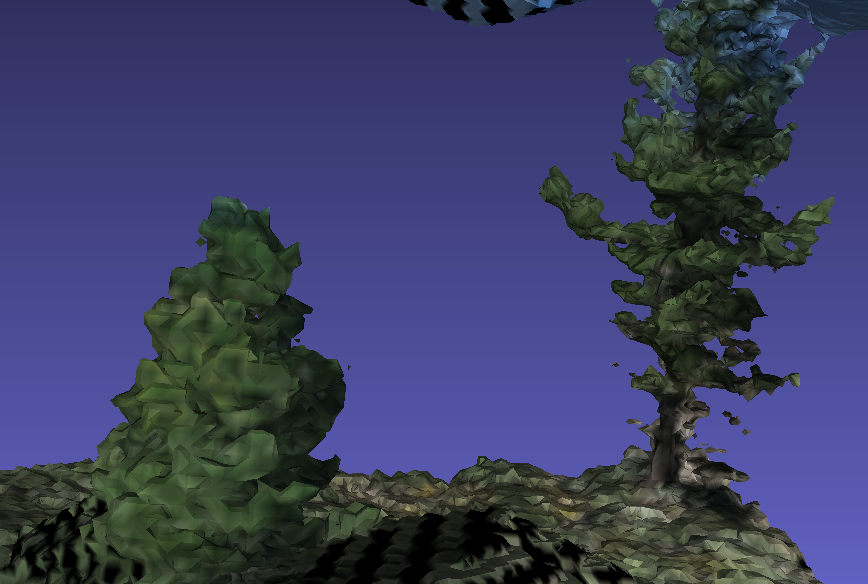}
   \caption{Mesh extracted from NeRF.}
   \label{fig:Ng2}
\end{subfigure}
\caption{\small A side by side of a 3D point-cloud and a 3D mesh generated via different reconstruction methods of the same imagery.}
\label{mesh}
\end{wrapfigure}

For this proof-of-concept work, we used an OFO video acquisition, collected using a DJI Phantom 4 Pro v2 drone, that included footage from many angles of an isolated lodgepole pine (\textit{Pinus contorta}) on the Tahoe National Forest near Graegle, California (39.67 deg W, 120.62 deg N). We processed the data in nerfstudio~\citep{tancik2023nerfstudio} and used their nerf-facto method to create a NeRF of the scene (Fig. \ref{nerf}). It is immediately obvious the significantly enhanced photo-realism of the NeRF compared to a SfM created mesh (Fig. \ref{fig:Ng1}). Detail such as branches, leaves, and trunks can be made out much more clearly which can aid in a variety of downstream tasks including species classification and trunk measurement. In Fig. \ref{fig:teaser}, we show another example of a NeRF, this time created from overhead imagery. As the images were taken from further overhead, less detail is visible, but it is still more photorealistic than the current 3D model in the OFO taken on a similar stand of trees with the same image collecting procedure.

The mesh extracted from the NeRF also has improved detail. A SfM point-cloud generated from the same data has a significant number of floaters. In Fig. \ref{mesh} we compare the two. The high number of artifacts prevents us from reasonably meshing the point-cloud. Those floaters can be removed through post-processing, but in general we are not seeing much detail in the individual trees, which is consistent with the SfM data already present in OFO. In addition to providing a richer visual experience, having more detail can serve a various scientific purposes. There are several qualitative tasks that typically require an expert walking through a forest, such as fuel load estimation or density verification, that could potentially be done virtually with a NeRF, saving significant travel time and costs as many forests are quite remote. NeRFs also allow the 3D qualitative experience to be saved, enabling more robust comparisons across time. 

\section{The Potential of NeRF in the OFO: Climate and Environmental Impacts}

Broad-extent forest inventory data is critical for informing management of forests in our era of changing climate, increasing drought stress, and unnaturally high-severity wildfires. Due to a century of intensive fire suppression and exclusion of indigenous forest stewardship through fire, dry forests in the western U.S. -- and many other areas around the globe -- have become unnaturally dense ~\citep{safford2017natural}. With more trees competing for the same finite pool of resources, particularly water, such ``overstocked" stands are at high risk of mortality due to drought~\citep{young2017long}. 
Forest mortality, whether through drought, fire, or other environmental stressors, has clear implications for carbon storage and myriad other ecosystem services. Understanding tree density is essential for proper forest management.  

Forest management such as mechanical thinning, prescribed fire, and reintroduction of beneficial wildfire can greatly ameliorate the stresses associated with unnaturally dense forest stands and improve forest resistance to drought and fire~\citep{young2020forest}, thus reducing the risk of catastrophic forest and carbon loss. However, resources for such forest management are stretched thin, and only a small fraction of forest area needing management each year receives treatment~\citep{north2021pyrosilviculture, north2015constraints}. Thus, data to inform efforts to prioritize forest management across space and time is critical. Because in-person assessments and surveys for all forests are not practical, modern tools to automate and virtualize forest measurement could thus provide substantial value to forest management efforts. 
The OFO aims to greatly increase the efficiency of such data collection using drone-based alternatives to traditional ground-based surveys and enable individuals such as forest managers to both collect and process the data into extensive forest inventory maps suitable for informing management decisions. The work presented here, and future expansions upon it, will help to improve the fidelity of drone-derived products to the real-world stand conditions, thus increasing their relevance for informing management to maintain healthy forests. 


\begin{wrapfigure}[12]{r}{0.4\textwidth}
\vspace{-1em}
\centering
\includegraphics[width=.3\textwidth]{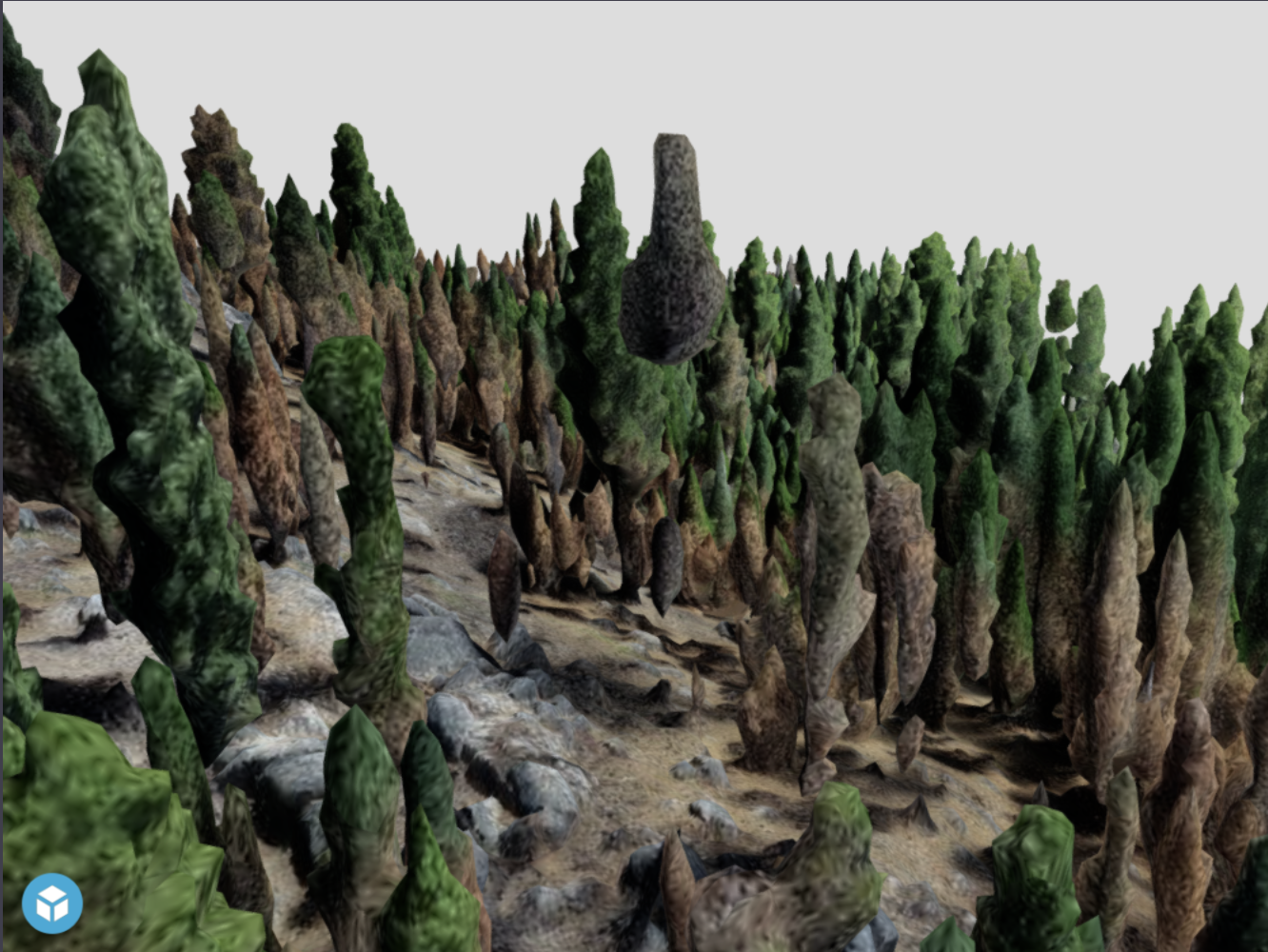}

\caption{A SfM reconstruction where trees are floating off the ground.}
\label{fig:sfmtrees}
\end{wrapfigure}

The vanilla NeRF proof-of-concept is just a gateway to exploring NeRF-based method for forestry. There are several variants of NeRF that could also improve the quality and types of data the OFO offers. The next immediate goal is to greatly scale up the size of the NeRF taking inspiration from other large-scale NeRF efforts~\citep{tancik2022block}. 
We also wish to do a better job filling in the hard-to-image understory: Trees often appear to be floating floating (Fig. \ref{fig:sfmtrees}). We plan to explore approaches that combine diffusion models with NeRF~\citep{deng2023nerdi, gu2023nerfdiff,yang2023learning}. We believe forests are particularly well suited to this approach as we have a strong prior that most everything we are imaging is a tree, and there are basic principles such as ``trees should be connected to the ground" that we can enforce to improve reconstruction quality. 

Another exciting extension of NeRF that could aid in forestry management is the incorporation of language models. LERF~\citep{kerr2023lerf} allows for natural-language searching through a NeRF, which if fine-tuned on species level forest data could be very useful for finding and counting species. As we see more multi-model models that can handle question answering, we could imagine merging those with NeRF as well to give more information about the 3D environment such as stand density, basal area, or biomass, all of which are important metrics in estimating carbon sequestration and fuel sources for wildfires. 

The OFO is compiling extensive imagery and ground truth data (in the form of geospatial locations, species labels, and diameter measurements) of forest stands across the western U.S. and ultimately around the world. We plan to curate a subset of the data in the form of well-defined ``challenge problems" to spur the development of new vision methods specifically designed for this high-impact application area. Already metrics for 3D reconstruction quality are a hotly debated topic in the community, as what constitutes an acceptable versus an unacceptable error is often dependent on the downstream application. In many forestry applications the key metric is faithful reconstruction of a tree's stem diameter at breast height (DBH), given it is the size metric around which the vast majority of forest ecology models and management prescriptions are based. With a specific downstream application in mind, along with extensive and high-quality ground-truth DBH data, we hope to provide a new application area and validation metric to the 3D reconstruction community. 

With this paper we invite the AI community to explore the OFO dataset for themselves and consider contributing methods to this vital application area. The introduction of NeRF has potential to not only improve the visual appearance of the 3D maps, but to allow for the introduction of a host of other AI-powered approaches that can greatly increase the types of secondary data produced. This data has huge potential to help in wildfire preparedness and estimating carbon sequestration, two objectives that become increasingly important as our climate changes. 

\subsubsection*{Acknowledgments}
This research used resources of the National Energy Research Scientific Computing Center (NERSC), a U.S. Department
of Energy Office of Science User Facility located at Lawrence Berkeley National Laboratory, operated under Contract No.
DE-AC02-05CH11231.

\bibliography{references}
\bibliographystyle{iclr2024_conference}

\end{document}

%% file: math_commands.tex
\usepackage{amsmath,amsfonts,bm}

\def\eqref#1{equation~\ref{#1}}

\def\1{\bm{1}}

\DeclareMathAlphabet{\mathsfit}{\encodingdefault}{\sfdefault}{m}{sl}
\SetMathAlphabet{\mathsfit}{bold}{\encodingdefault}{\sfdefault}{bx}{n}

